\documentclass[amsmath,amssymb,amsfonts,pra,aps,floatfix,11pt,tightenlines,raggedbottom,superscriptaddress]{revtex4-2}

\usepackage[svgnames,x11names,table]{xcolor}
\usepackage{hyperref}
\usepackage[protrusion=true,expansion=true]{microtype} 
\usepackage{graphicx}%
\usepackage{multirow}%
\usepackage{amsthm}%
\usepackage{mathrsfs}%
\usepackage[title]{appendix}%
\usepackage{xcolor}%
\usepackage{textcomp}%
\usepackage{booktabs}%
\usepackage{listings}%

\hypersetup{
    colorlinks=true,
    urlcolor=MediumBlue,
    linkcolor=MediumBlue,
    citecolor=MediumBlue,
    filecolor=MediumBlue,
    allcolors=MediumBlue,
    pdftitle={Neuromorphic Visual Scene Understanding with Resonator Networks},
    pdfsubject={cs.NC},
    pdfauthor={Alpha Renner},
    pdfkeywords={Hyperdimensional Computing, Vector Symbolic Architecture (VSA), Neuromorphic Hardware, Spiking Neural Networks (SNN), Loihi},
}

\begin{document}

\title[Neuromorphic Visual Scene Understanding with Resonator Networks]{Neuromorphic Visual Scene Understanding with Resonator Networks}

\author{Alpha Renner} \email{alpren@ini.uzh.ch}
 \affiliation{Institute of Neuroinformatics, University of Zurich and ETH Zurich, Switzerland}
 \affiliation{Forschungszentrum Jülich, Germany}
\author{Lazar Supic}
\affiliation{Accenture Labs, San Francisco, CA, USA}
\author{Andreea Danielescu}
\affiliation{Accenture Labs, San Francisco, CA, USA}
\author{Giacomo Indiveri}
 \affiliation{Institute of Neuroinformatics, University of Zurich and ETH Zurich, Switzerland}
\author{Bruno A. Olshausen}
\affiliation{Redwood Center for Theoretical Neuroscience, UC Berkeley, CA, USA}
\author{Yulia Sandamirskaya}\email{yulia.sandamirskaya@zhaw.ch}
\affiliation{Intel Labs, Zürich, Switzerland}
\affiliation{ZHAW Zurich University of Applied Sciences, Wädenswil, Switzerland}
\author{Friedrich T. Sommer}\email{fsommer@berkeley.edu}
\affiliation{Redwood Center for Theoretical Neuroscience, UC Berkeley, CA, USA}
\affiliation{Intel Neuromorphic Computing Lab, Intel Labs, Santa Clara, CA, USA}
\author{E. Paxon Frady} \email{e.paxon.frady@intel.com}
\affiliation{Intel Neuromorphic Computing Lab, Intel Labs, Santa Clara, CA, USA}
















\begin{abstract}
Analyzing a visual scene by inferring the configuration of a generative model is widely considered the most flexible and generalizable approach to scene understanding. Yet, one major problem is the computational challenge of the inference procedure, involving a combinatorial search across object identities and poses.
Here we propose a neuromorphic solution exploiting three key concepts: (1) a computational framework based on Vector Symbolic Architectures (VSA) with complex-valued vectors; (2) the design of Hierarchical Resonator Networks (HRN) to factorize the non-commutative transforms translation and rotation in visual scenes; (3) the design of a multi-compartment spiking phasor neuron model for implementing complex-valued resonator networks on neuromorphic hardware.
The VSA framework uses vector binding operations to form a generative image model in which binding acts as the equivariant operation for geometric transformations. A scene can, therefore, be described as a sum of vector products, which can then be efficiently factorized by a resonator network to infer objects and their poses.
The HRN features a partitioned architecture in which vector binding is equivariant for horizontal and vertical translation within one partition and for rotation and scaling within the other partition. The spiking neuron model allows mapping the resonator network onto efficient and low-power neuromorphic hardware.
Our approach is demonstrated on synthetic scenes composed of simple 2D shapes undergoing rigid geometric transformations and color changes. A companion paper demonstrates the same approach in real-world application scenarios for machine vision and robotics.

\end{abstract}

\flushbottom
\maketitle
\thispagestyle{empty}

Visual scene understanding is a long-standing problem of machine vision and artificial intelligence.  
The disentanglement of scene objects into their individual properties is promising but also a notoriously hard --and largely unsolved-- computational problem because it requires searching over a very large space of possible configurations of how objects can be combined with variations in pose, color, lighting, and other features \citep{poggio1987computational,yildirim2020efficient, williams2023structured}.
The use of convolutional neural networks (CNN) has been proposed, an approach that typically requires large amounts of training data and additional augmentations to handle variations in pose or style. The resulting performance is often brittle~\citep{szegedy2013intriguing,madry2017towards} and easily fooled~\citep{nguyen2015deep,kurakin2018adversarial}. Further, the operation of CNNs is opaque with the scene information entangled in their parameters, which makes it difficult to trace information flow and to fix the failure modes.

It has long been proposed that the brain solves visual scene understanding via ``analysis-by-synthesis'' whereby a generative model is used to infer the components of a scene that best explain the visual input~\citep{mackay1956towards,neisser-book,Yuille_Kersten06}.
However, this type of inference incurs a high computational cost, which has prevented the widespread deployment of this strategy.  
Recent work has shown that for workloads that require recurrent iterative computations, like inference in generative models, \textit{neuromorphic computing} can vastly outperform CPU and GPU-based approaches~\citep{davies2021advancing}. 
Specifically, custom spike-based neuromorphic chips~\citep{TrueNorth14,Furber_etal14,Moradi_etal18,Pei_etal19,davies2021advancing} accelerate computing times and reduce power consumption thanks to their parallelism, in-memory processing~\citep{Indiveri_Liu15}, sparsity, and event-based~\citep{gallego_2020a} nature. 

Our neuromorphic approach to scene analysis employs a programming framework from Cognitive Science that represents information as high-dimensional vectors and then computes on these representations via an explicit algebra~\citep{PlateHolographic1995,kanerva_1996,GaylerAnalogical1998}.
The framework, known as Vector Symbolic Architectures (VSAs)~\citep{GaylerJackendoff2003}, or Hyperdimensional Computing (HC)~\citep{kanerva_2009}, offers an explicit binding operation that addresses the so-called feature binding problem in conventional artificial neural networks~\citep{Malsburg81,von1995binding,feldman_2012}.
Here, we leverage recent developments in VSA for designing a neuromorphic algorithm~\citep{KleykoComputingParadigm2021} for scene analysis: 1) a mathematical framework that extends VSAs to represent continuous variables and functions~\citep{frady2021VFA}, and 2) a \emph{resonator network} that efficiently solves multi-factor vector factorization in VSAs~\citep{frady_2020_resonator,kent_2020resonator}.  The first development enables us to encode an image in a VSA representation such that binding acts as the equivariant operation for specific geometric transformations~\citep{frady2021VFA}, while the second one makes it tractable to infer objects and their transformations via vector factorization~\citep{frady_2020_resonator,kent_2020resonator}.

The proposed approach falls within the larger family of multilinear models for inferring object shapes and their transformations in the context of a generative image model.  These include early proposals by Pitts \& McCulloch (1947)\citep{pitts1947we} and Hinton (1981)\citep{hinton1981parallel} for remapping sensory information into a canonical reference frame, neurobiological models such as dynamic routing~\citep{olshausen1993neurobiological}, map-seeking circuits \citep{arathorn2002map,arathorn2004computation}, as well as bilinear models that learn to disentangle form vs. motion (or `style' vs. `content')~\citep{tenenbaum1996separating,freeman1997learning,vasilescu2002multilinear,olshausen2007bilinear,chau2020disentangling}.

Here, we first describe how an image can be encoded in a VSA representation so that the binding operation is the equivariant operation for translation. With the same encoding scheme, we then formulate a generative model of a scene composed of translated template shapes and show how resonator networks \citep{frady_2020_resonator} can infer translations and object templates that generated a given image.
Extending this approach, we develop an algorithm employing a new \emph{hierarchical resonator network} for analyzing scenes composed of arbitrary rigid transforms of shape templates.
Finally, we demonstrate how to implement the essential components of the hierarchical resonator network on Intel’s neuromorphic research chip, Loihi~\citep{davies2018loihi}, utilizing an efficient spike-timing code.

\section*{Representing images with hypervectors}

High-dimensional random vectors are approximately orthogonal; that is, pairs of vectors are very likely to have a small inner product. VSAs leverage this separation by representing individual symbols with random vectors ($\mathbf{a}, \mathbf{b}$, etc.) in an $N$-dimensional space \citep{PlateHolographic1995,kanerva_2009}. VSAs typically offer two complementary dyadic vector operations to form composite data structures, preserving the vector dimension: superposition and binding. 
The $N$-dimensional vectors representing atomic and composite data structures during a computation are called {\em hypervectors}. 
Recently, VSAs have been generalized to Vector Function Architectures (VFAs) \citep{frady2021VFA} that can represent in hypervectors not only data structures of discrete symbols but also real-valued quantities \citep{frady_2019precession, komer_2019} and functions \citep{frady2021VFA}.
During computation in VFAs, the execution of vector operations is interleaved with parsing/decoding and error correction, exploiting similarity-based access of the interpretable hypervectors stored in the so-called \emph{codebook}, for example, through nearest-neighbor search or auto-associative content-addressable memory
~\citep{KleykoSurveyVSA2021Part1,KleykoSurveyVSA2021Part2}. 

Here, we use a VFA built on Fourier Holographic Reduced Representations (FHRR)~\citep{PlateNested1994, PlateHolographic1995}, a VSA whose atomic hypervectors are composed of phasors, i.e., complex-valued variables with unit amplitude.
Similarity between two FHRR hypervectors is measured by the real part of the normalized inner product, $\frac{1}{N} \Re (\mathbf{a}^\dagger \mathbf{b})$, where $\mathbf{a}^{\dagger}$ is the complex conjugate vector transpose. 
The binding operation in FHRR is the Hadamard product or element-wise multiplication $\odot$, with $\mathbf{a} \odot \mathbf{b}$ for \emph{binding}, and $\mathbf{a} \odot \mathbf{b}^*$ for \emph{unbinding}, where $\mathbf{b}^*$ is the complex conjugate. The \emph{superposition} operation is vector summation, $\mathbf{a} + \mathbf{b}$.

To encode an image as a hypervector, VFA index vectors are created to encode pixel location. 
We choose two fixed complex-valued FHRR vectors $\mathbf{h}$ and $\mathbf{v}$, for horizontal and vertical position respectively, whose elements are complex phasors of unit amplitude (i.e., $h_j=e^{\imath \varphi_j}$) and randomly assigned phase ($\varphi_j \sim \mathcal{U}[0, 2\pi]$, with $\mathcal{U}[0, 2\pi]$ the uniform distribution).
A pixel at the Cartesian image coordinates $x$, $y$ is then represented by the index vector $ \mathbf{h} ^{x} \odot \mathbf{v} ^{y}$. Exponentiating vectors $\mathbf{h}$ and $\mathbf{v}$ essentially spins the phase of each element $j$ proportional to the value of $\varphi_j$.

Following \citep{frady2021VFA}, the image $Im(x,y)$ is encoded as a function over the pixel space via the superposition of index vectors weighted by their corresponding image pixel values $\mathbf{s} = \sum_{x,y} Im(x,y) \cdot \mathbf{h}^{x} \odot \mathbf{v}^{y}$.
This encoding is similar to the Discrete Fourier transform but with frequencies chosen randomly~\citep{rahimi2007random} instead of regular spacing. 
Interestingly, important properties of the Fourier transform, such as the convolution theorem, remain valid even with the frequencies randomized.
Repeated binding ($\mathbf{h}^{x} $) is currently also commonly used for position encoding in transformers~\citep{su2024roformer}.
Further, three random vectors index the color channels red, green, and blue, $\mathbf{G} = [ \mathbf{r}, \mathbf{g}, \mathbf{b} ] \in \mathbb{C}^{N \times 3}$. The hypervector representation of a color image is then given as:
\begin{equation}
    \mathbf{s} = \sum_{x,y,c} Im(x, y, c) \cdot \mathbf{G}_{c} \odot \mathbf{h} ^{x} \odot \mathbf{v} ^{y} =: \mathbf{\Phi} \; \mathbf{I},
    \label{eq:encode}
\end{equation}
where $\mathbf{G}_{c}$ indicates the vector representing a color channel.
The definition on the right of (\ref{eq:encode}) makes it explicit that hypervector encoding of an image is a linear projection, with $\mathbf{I} \in \mathbb{R}^{(3P_xP_y)}$ the image reshaped as a vector, and $\mathbf{\Phi}\in \mathbb{C}^{N \times (3P_xP_y)}$ the codebook matrix of hypervectors for each index configuration $\{x,y,c\}$,  where $P_x, P_y$ are the dimensions of the image in pixels. 
Conversely, decoding the image from the hypervector uses the conjugate transpose as the linear transform, $\mathbf{I} = \frac{1}{N}\Re (\mathbf{\Phi}^\dagger \mathbf{s})$. 
Since the codebook entries are only approximately orthogonal, decoding introduces small amounts of noise in the image reconstruction, which can be quantified and mitigated \citep{frady_2018sequence}. In this context, these noise effects are minimal. 

Importantly, image encoding with (\ref{eq:encode}) enables binding to act as the \emph{equivariant vector operation} for image translation:
\begin{equation}
    \mathbf{s} \odot \mathbf{h}^{\Delta x} \odot \mathbf{v}^{\Delta y} = \sum_{x, y} Im(x, y) \cdot \mathbf{h}^{x + \Delta x} \odot \mathbf{v} ^ {y + \Delta y} = \sum_{x,y} Im(x - \Delta x, y - \Delta y) \cdot \mathbf{h}^x \odot \mathbf{v}^y.
    \label{imagetrans}
\end{equation}
Also, note that image translation is well-defined for continuous values of $\Delta x$, $\Delta y$, allowing the recognition of shapes shifted by fractions of a pixel.

To facilitate understanding of the notation and symbols here and in the following sections, we provide Supplementary Table \ref{tab:notation_table} as a reference.

\section*{A generative model of scenes using VSA vector operations}

We demonstrate scene understanding on simple synthetic images composed of object templates, in our case, letters, that are translated and given one of 7 colors. 
The scene understanding task is to extract the identities, colors, and locations of objects from an input image. 

In the VFA framework, the generative model for the synthetic images can be formulated as follows.
The set of the image templates of the letters are aligned (see Methods) and written as the matrix $\mathbf{P} \in \mathbb{R} ^ {(P_x P_y) \times D}$,  where $D=26$ is the number of different templates. 
As in (\ref{eq:encode}), a letter template is represented by the hypervector
\begin{equation}
    \mathbf{d}_a = \sum_{x,y} \mathbf{P}_a (x, y) \cdot \mathbf{h}^{x} \odot \mathbf{v}^{y}.
    \label{eqn:vfa_im}
\end{equation}
with $\mathbf{P}_a (x, y)$ the pixel intensity value of a template image indexed by a
at position $(x,y)$.
Each hypervector for the templates is stored in the matrix $\mathbf{D} = \mathbf{\Phi}_\mathbf{P} \mathbf{P}$, with $\mathbf{\Phi}_\mathbf{P}$ containing the vectors for each pixel as described in (\ref{eqn:vfa_im}).

\begin{figure}[t]
    \centering
    \includegraphics[width=1\textwidth]{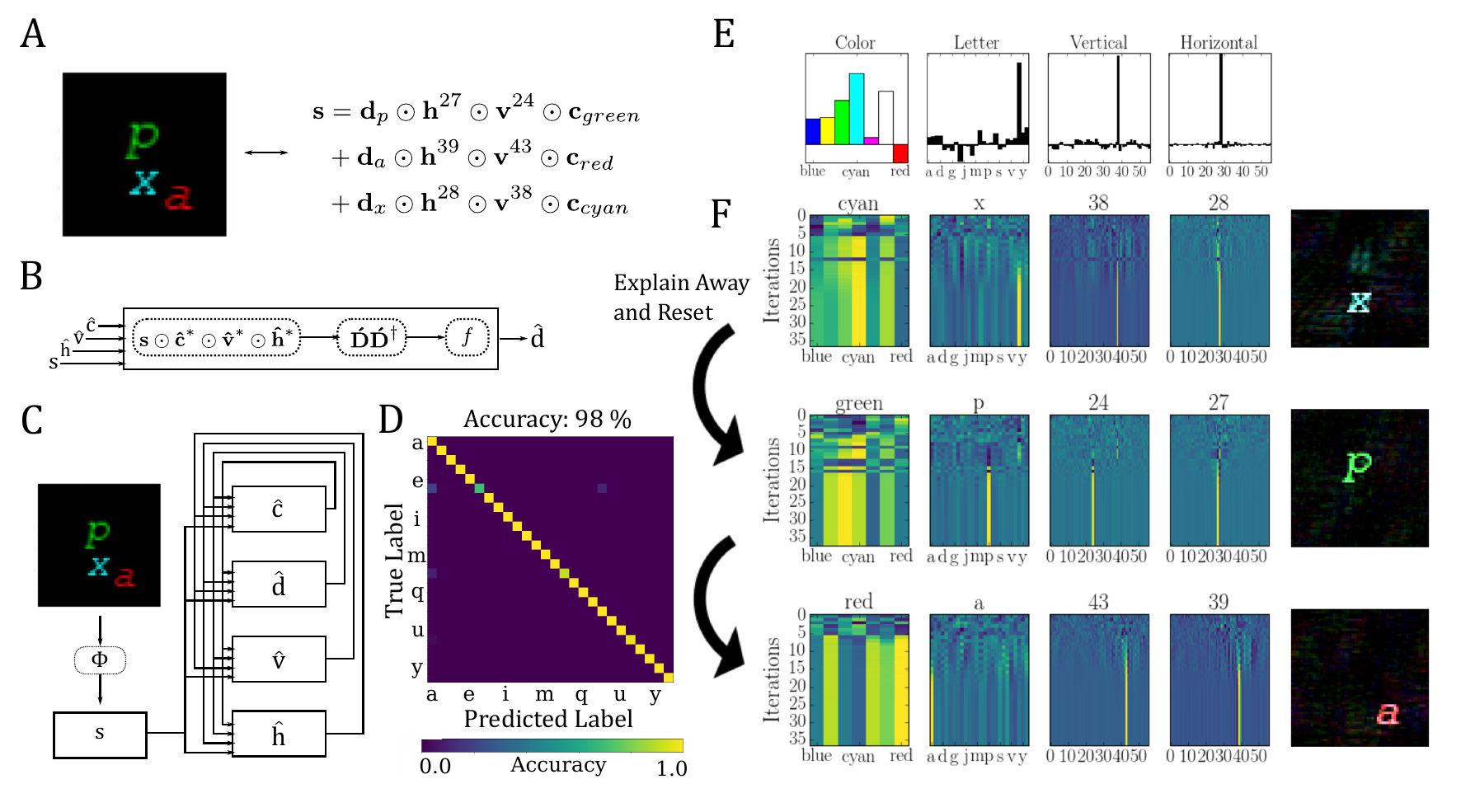}
    \caption{Resonator network for inferring shape, color, and translation. \textbf{A.} A synthetic scene and the generative VSA representation.
    \textbf{B.} A resonator module. 
    \textbf{C.} Encoding and communication in the resonator network.
    \textbf{D.} Confusion matrix on translation benchmark task with a single object. The overall performance of the network is 98.4\%.
    \textbf{E.} The weighted factor estimates in each resonator module. The maximum value is taken as the output. 
    \textbf{F.} The four dynamic estimates in the resonator network are each visualized as a heatmap, with time represented vertically and each component represented horizontally. After several iterations, the resonator network converges to a solution and remains stable (first row corresponds to panel E). The decoded output is visualized to the right of each row. The object is then `explained away'. The resonator network is reset and converges to another solution, which describes a different object in the scene (rows 2 and 3). 
}
    \label{fig:res_circuit}
\end{figure}

The equivariant vector binding operation is used to represent shape templates with specific positions and colors. Vector superposition is used to compose scenes from multiple objects. 
All told, the VFA representation of a generated scene composed of $L$ objects is: 
\begin{equation}
    \mathbf{s} = \sum_{i=1}^L \mathbf{d}_{p_i} \odot \mathbf{h} ^ {x_i} \odot \mathbf{v} ^ {y_i} \odot \mathbf{c}_{c_i}.
    \label{eq:scene_tr}
\end{equation}
In the generative model of artificial scenes based on (\ref{eq:scene_tr}), each factor of variation ($p_i, x_i, y_i, c_i$) is sampled uniformly. One of 7 colors is chosen, given by a matrix $\mathbf{B} \in \mathbb{R}^{3 \times 7}$ with, for instance, $\mathbf{B}_{cyan} = [0, 1, 1]$. The VSA codebook for colors is $\mathbf{C} = \mathbf{G} \mathbf{B}$. 
For an example scene and its corresponding hypervector (\ref{eq:scene_tr}),  see Fig.~\ref{fig:res_circuit}A.

To enable inference in the generative model (\ref{eq:scene_tr}), each generative factor has to be represented by sufficiently dissimilar hypervectors, including elements with correlations in the image domain, like ``c'' versus ``o''.
Decorrelated hypervectors for overlapping shape templates are produced by whitening the generative factors using singular value decomposition \citep{tenenbaum2000separating}, $\mathbf{P} = \mathbf{U} \mathbf{\Sigma} \mathbf{V}$. The whitened templates $\mathbf{\acute{P}} = \mathbf{U} \mathbf{V}$ are encoded into hypervectors similar to (\ref{eqn:vfa_im}) and stored in the codebook $\mathbf{\acute{D}} = \mathbf{\Phi_P} \mathbf{\acute{P}}$. 
A similar whitening procedure is used to generate the codebook for color, $\mathbf{\acute{C}} \in \mathbb{C}^{N \times 7}$.

\section*{Inference with the resonator network}

A generative model (\ref{eq:scene_tr}) enables, in principle, understanding of a given input image by inferring the causal factors that best explain the data. However, inference in generative models is computationally expensive \citep{teh2003energy} as it involves a potentially exhaustive search across all templates in all possible poses. The VSA formulation (\ref{eq:scene_tr}) permits fast parallel implementations of this search.
A given image to be analyzed is first transformed via equation (\ref{eq:encode}) into a hypervector $\mathbf{s}$. 
Inference of the image content then involves fitting the input vector $\mathbf{s}$ by the best matching templates and transforms contained in the model. 
In particular, each term of the sum in (\ref{eq:scene_tr}) represents one image component, and inferring its properties requires factorizing the corresponding term into specific hypervectors.

The problem of vector factorization is common in VSA algorithms, and \emph{resonator networks} can solve it efficiently \citep{frady_2020_resonator, kent_2020resonator}. A resonator network is composed of modules, one for each factor of variation in the generative model, that are recurrently interconnected.

A resonator network module contains three stages: a VSA binding stage, a linear auto-associative memory for code vectors \citep{kohonen1974adaptive} representing one factor, and an element-wise saturation function or normalization (Fig. \ref{fig:res_circuit}B). 
The resonator network (\ref{eqn:resonator_transla}) solves the inference problem dynamically. Starting from random seeds, in each iteration, a module decodes the estimate of its own factor from $\mathbf{s}$ by unbinding the estimates from all other modules. The decoded vector is then compared to the module's codebook. 
Based on vector similarity, the auto-associative memory \emph{cleans up} the decoded vector to resemble one or a superposition of valid codebook vectors. After applying the transfer function $f$, this new estimate is sent to the other modules.
The complete dynamic equations of the resonator network for inference in (\ref{eq:scene_tr}) are: 
\begin{align}
\begin{split}
\mathbf{\hat{c}}(t+1) &= f \left( \mathbf{\acute{C}}\mathbf{\acute{C}}^\dagger \left( \mathbf{s} \odot \mathbf{\hat{d}}^*(t) \odot \mathbf{\hat{v}}^*(t) \odot \mathbf{\hat{h}}^*(t) \right) \right), \\
\mathbf{\hat{d}}(t+1) &= f \left( \mathbf{\acute{D}}\mathbf{\acute{D}}^\dagger \left( \mathbf{s} \odot \mathbf{\hat{c}}^*(t) \odot \mathbf{\hat{v}}^*(t) \odot \mathbf{\hat{h}}^*(t) \right) \right), \\
\mathbf{\hat{v}}(t+1) &= f \left( \mathbf{V}\mathbf{V}^\dagger \left( \mathbf{s} \odot \mathbf{\hat{d}}^*(t) \odot \mathbf{\hat{c}}^*(t) \odot \mathbf{\hat{h}}^*(t) \right) \right), \\
\mathbf{\hat{h}}(t+1) &= f \left( \mathbf{H}\mathbf{H}^\dagger \left( \mathbf{s} \odot \mathbf{\hat{d}}^*(t) \odot \mathbf{\hat{v}}^*(t) \odot \mathbf{\hat{c}}^*(t) \right) \right), \\
\end{split}
\label{eqn:resonator_transla}
\end{align}
with $f(\mathbf{x})_i = x_i/|x_i|$ (phasor projection) or, alternatively, $f(\mathbf{x})_i = x_i/||\mathbf{x}||_2$ (normalization) (see Methods).  $\mathbf{V}$, $\mathbf{H}$ are codebooks representing vertical and horizontal pixel coordinates, and $\mathbf{\acute{C}}$, $\mathbf{\acute{D}}$ are codebooks representing color and object shape as described above. 

The dynamics of equation (\ref{eqn:resonator_transla}) successively improve the joint estimate of all factors (Fig. \ref{fig:res_circuit}C). 
Importantly, individual modules do not settle immediately at a single estimate for their factor. In early iteration steps, they produce a superposition of many possible factors, which enables parallel search through the combinatorics of solutions. In later iterations, the interaction between modules narrows the search down to a single estimate of the identity, pose, and color of one scene component, and the network converges (Fig. \ref{fig:res_circuit}D, E).

For parsing scenes with multiple components, the analysis process is repeated after previously identified image components are subtracted from $\mathbf{s}$ as part of an outer loop. This subtraction is akin to ``explaining away'' or ``deflation'' \citep{burden2015numerical}.
Alternatively, analysis of multiple objects can also be done with several instances of the resonator running in parallel, the multi-headed resonator (see Supplementary Information and Supplementary Fig. \ref{fig:multihead}). 

To evaluate the model performance, we designed benchmark tasks for invariant letter recognition in composed scenes. Fig. \ref{fig:res_circuit}F shows accuracy results for classifying a single letter in a scene (see Methods). For the recognition task with all 26 letters, the network was $98\%$ accurate at identifying letter templates regardless of color or translation ($N=10,000$); see Supplementary Information and Supplementary Fig. \ref{fig:multiletter} for the analysis of scenes with multiple letters. 

\section*{Scenes with rotated and scaled objects}

\begin{figure}[ht]
    \centering
    \includegraphics[width=0.95\textwidth]{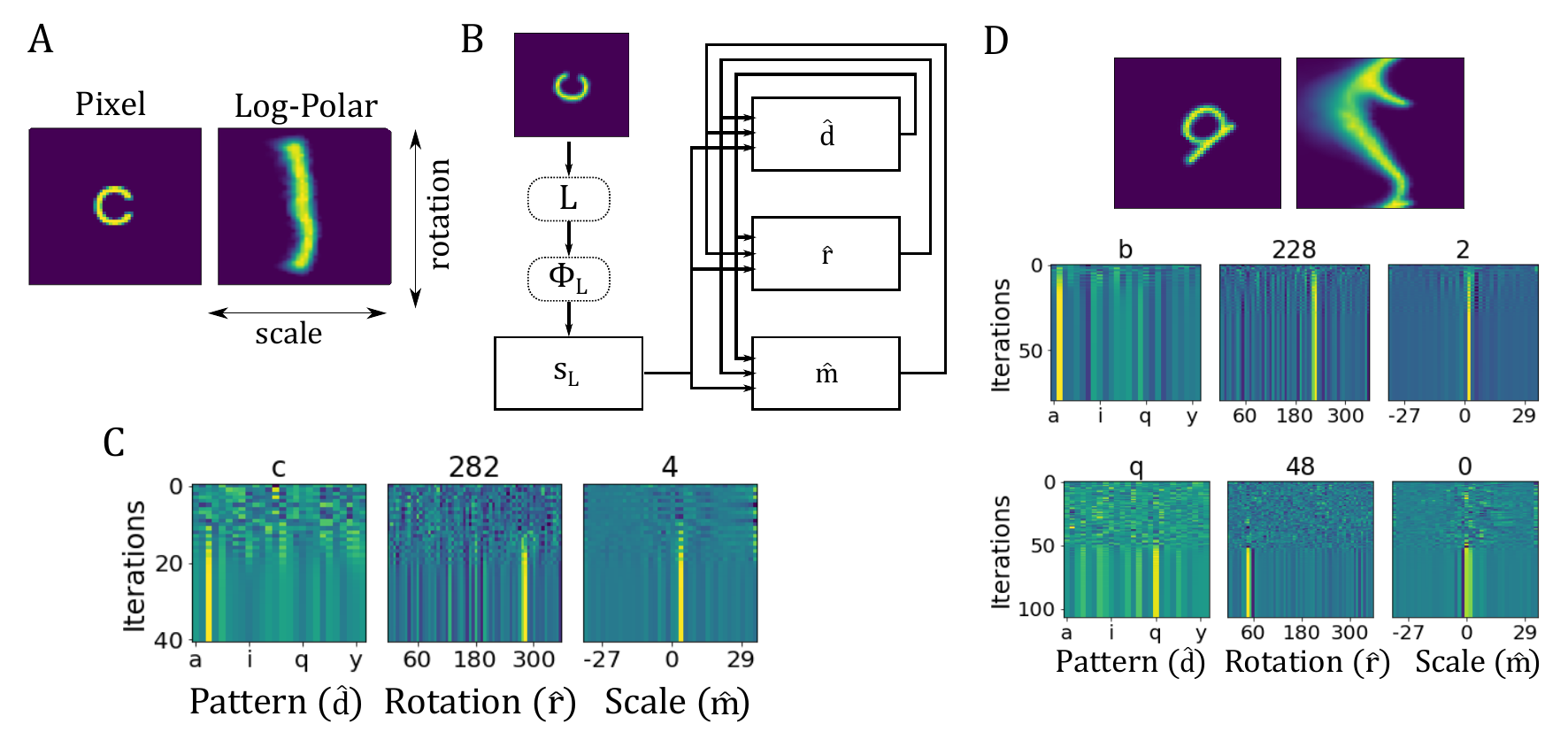} 
    \caption{Resonator network for rotation and scale. \textbf{A.} Translation in log-polar space results in rotation and scaling in Cartesian space. 
    \textbf{B.} Diagram of resonator network for inferring shape, rotation, and scaling of input images. 
    \textbf{C.} Example of network dynamics. 
    \textbf{D.} Symmetries of the template lead to ambiguous factorizations. Two examples are shown with different random initializations. The resonator network will converge to one of the ambiguous factorizations (letters `b' or `q').}
    \label{fig:rot_scale}
\end{figure}

The approach can also be applied to scenes where a centered object is transformed by rotation and scaling. We use the fact that rotation and scaling in Cartesian space becomes translation in log-polar space (Fig.~\ref{fig:rot_scale} A), with $\mathbf{L} \in \mathbb{R} ^ {L_m L_r \times P_x P_y}$ the log-polar transform matrix and $L_m, L_r$
the pixel dimensions in log-polar space.  Thus, in log-polar coordinates, the binding operation becomes the equivariant transform for rotation and scaling. From a generative model of rotated and scaled object templates in log-polar coordinates, we can construct the following resonator network (Fig.~\ref{fig:rot_scale} B):
\begin{align}
\begin{split}
\mathbf{\hat{d}}(t+1) &= f \left( \mathbf{\acute{D}_L}\mathbf{\acute{D}_L}^\dagger \left( \mathbf{s_L} \odot \mathbf{\hat{r}}^*(t) \odot \mathbf{\hat{m}}^*(t) \right) \right), \\
\mathbf{\hat{r}}(t+1) &= f \left( \mathbf{R}\mathbf{R}^\dagger \left( \mathbf{s_L} \odot \mathbf{\hat{d}}^*(t) \odot \mathbf{\hat{m}}^*(t) \right) \right), \\
\mathbf{\hat{m}}(t+1) &= f \left( \mathbf{M}\mathbf{M}^\dagger \left( \mathbf{s_L} \odot \mathbf{\hat{d}}^*(t) \odot \mathbf{\hat{r}}^*(t) \right) \right). \\
\end{split}
\label{eqn:lp_resonator}
\end{align}
Here, the codebooks $\mathbf{R}$ and $\mathbf{M}$ contain the vector symbols for each log-polar coordinate, e.g. $\mathbf{R} = [\mathbf{r}^1, \mathbf{r}^2, ... , \mathbf{r}^{L_r}]$, $\mathbf{M} = [\mathbf{m}^1, \mathbf{m}^2, ..., \mathbf{m}^{L_m}$]. The index vector $\mathbf{r}$ is designed to obey periodic boundary conditions, such that translated pixels wrap around the image \citep{frady2021VFA}. The codebook $\mathbf{\Phi_L} \in \mathbb{C}^{N \times L_m L_r}$ contains the binding products of rotation and scale hypervectors, similar to $\mathbf{\Phi}$ (\ref{eqn:vfa_im}). 
Further, the codebook $\mathbf{\acute{D}_L}=\mathbf{\Phi_L} \mathbf{L} \mathbf{\acute{P}}$ contains the whitened letter patterns in the log-polar space. 

The example of rotation and scale invariant inference with the resonator network (\ref{eqn:lp_resonator}) in Figure~\ref{fig:rot_scale} highlights the general problem that image components can have more than one valid explanation because of symmetries, for example, `b' versus `q.' 
For such ambiguous inputs, the resonator network offers one interpretation, depending on the random initialization and other noise sources (Fig.~\ref{fig:rot_scale}D), but will not indicate the existence of alternative interpretations. 

\section*{Scenes with rigid, non-commutative transforms}

\begin{figure}[ht]
    \centering
    \includegraphics[width=\textwidth]{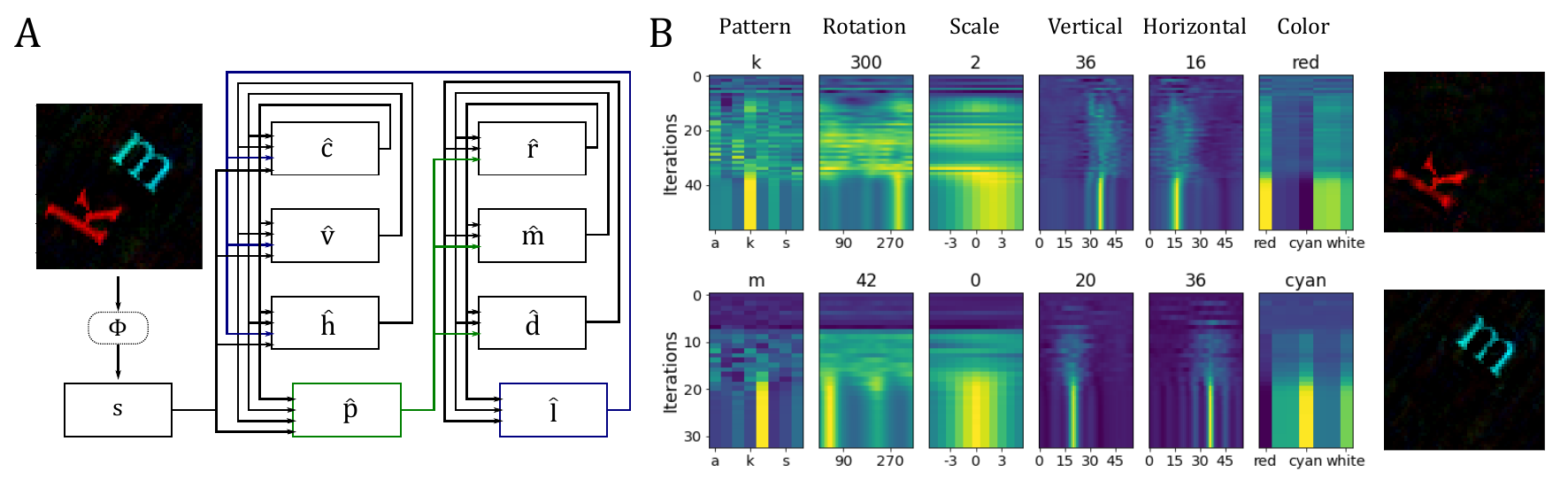}
    \caption{The hierarchical resonator network for inferring rigid transforms. \textbf{A.} Schematic diagram of the hierarchical resonator network. 
    \textbf{B.} The dynamics of the resonator network identifying objects in the input scene.}
    \label{fig:hier_res}
\end{figure}

The next step toward analyzing realistic scenes is the ability to identify object templates transformed by arbitrary rigid transforms, composed of translation, rotation, scale, and color. Building on the two previous models, the corresponding generative model of scenes has to include a log-polar transform matrix in the high-dimensional VSA space $\mathbf{\Lambda} = \mathbf{\Phi_L} \mathbf{L} \mathbf{\Phi^\dagger}$.
The generative model of an image synthesized from rigid transforms of shape templates can then be written as:
\begin{equation}
    \mathbf{s} = \sum_i \mathbf{c}_{c_i} \odot \mathbf{h}^{x_i} \odot \mathbf{v} ^ {y_i} \odot \mathbf{\Lambda}^{-1} (\mathbf{r} ^ {r_i} \odot \mathbf{m} ^ {m_i} \odot \mathbf{d}_{p_i} ), 
    \label{eq:gen_synth}
\end{equation}
The inference in this generative model can be performed in an adequately designed resonator network. Corresponding to the factors in (\ref{eq:gen_synth}), the network consists of six fully connected factor modules that all require coordinate transforms, $\mathbf{\Lambda}$ or $\mathbf{\Lambda}^{-1}$, in their binding stages, for full equations see Methods (\ref{eq:hierarch_res}). 
As depicted in Figure~\ref{fig:hier_res}A, the network is bisected into two partitions: one using Cartesian and one log-polar coordinates. Each partition has one additional module that serves as the communication bridge to the other partition. 
Conveniently, the bridge modules have the same internal stages as other resonator modules:
\begin{eqnarray}    
\mathbf{\hat{l}}(t+1) &=& \mathbf{\Lambda}^{-1} \big( \mathbf{\hat{r}}(t) \odot \mathbf{\hat{m}}(t) \odot \mathbf{\hat{d}}(t) \big),\label{eq:bridgeeq1}\\
\mathbf{\hat{p}}(t+1) &=& \mathbf{\Lambda} \big( \mathbf{s} \odot \mathbf{\hat{c}}^*(t) \odot \mathbf{\hat{h}}^*(t) \odot \mathbf{\hat{v}}^*(t)  \big). 
\label{eq:bridgeeq2}
\end{eqnarray}

\begin{figure}[ht]
    \centering
    \includegraphics[width=1.0\textwidth]{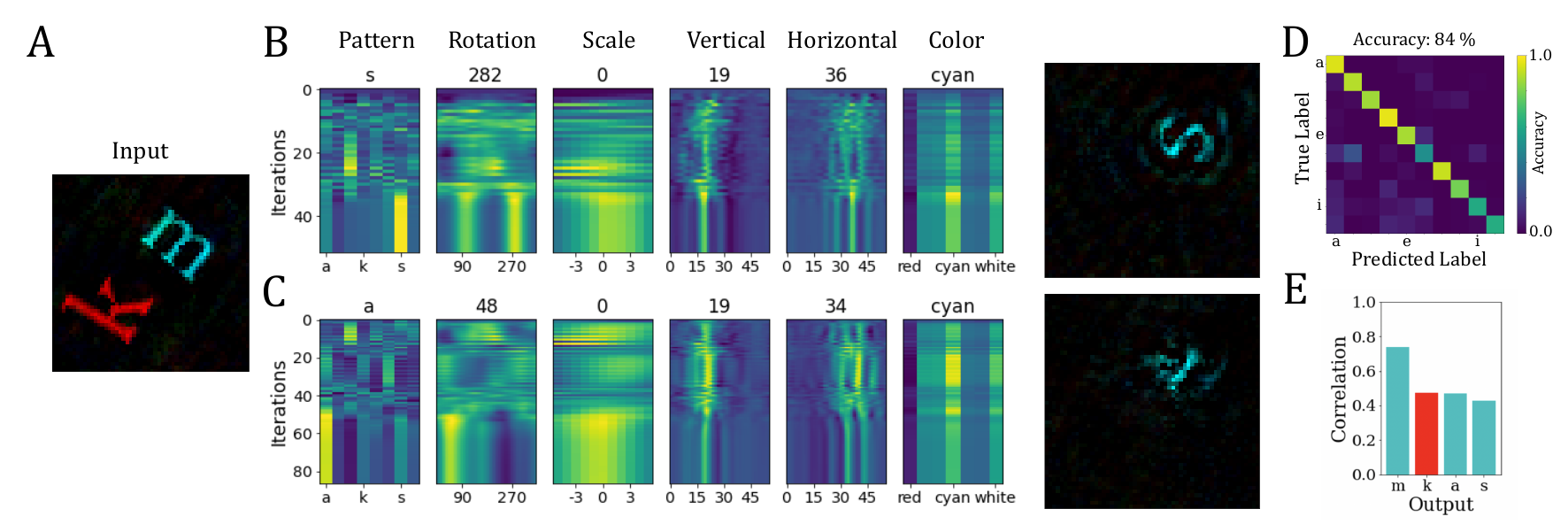}
    \caption{Local minima in the hierarchical resonator network. \textbf{A}. Input image. \textbf{B, C}. Two incorrect runs of the network are visualized. 
    \textbf{D}. Confusion matrix of object classes. 
    \textbf{E}. Correlations between the incorrect explanations and the input are on par with the correct explanations (right panel).}
    \label{fig:hier_res_err}
\end{figure}

A successful example of inference with the hierarchical resonator network is shown in Figure~\ref{fig:hier_res}B. The upper row shows a factorization process, revealing the letter ``k'', and the lower row shows a second factorization process, revealing the letter ``m''. Note how the estimates of all factors are undecided and blurry in early iteration steps and become sharp quite suddenly during iteration.  
Conversely, Figure~\ref{fig:hier_res_err} shows an unsuccessful inference example. In this case, the first factorization falsely explains the arched portion of the ``m'' with a rotated ``s''. After subtracting the ``s'', there are still parts of the ``m'' left, which the second factorization run falsely explains as an ``a''. 

We performed another set of benchmark experiments for rigid transforms. The task was reduced to a subset of 10 letters and about half ($\pm89^\circ$) of the rotations in order to reduce the complexity and the number of ambiguous scenes, as many of the ambiguities are due to $180^\circ$ rotational symmetry. 
In our benchmark experiment, the system identifies the correct letter with an accuracy of 84\% (Fig. \ref{fig:hier_res_err}D).

When the network fails, it often converges to a spurious solution composed of a complex mixture of generative factors. 
Even though these examples did not recover the true generative factors, the spurious solutions still have notable correlations with the input (Fig. \ref{fig:hier_res_err}E). 
In the Methods, we describe some modifications to the resonator network dynamics that mitigate these errors, such as non-linearities that encourage sparse solutions. However, these modifications cannot completely eliminate this issue, which has also been reported in other generative model approaches for scene analysis \citep{arathorn2002map}.

\section*{Neuromorphic implementation using spike times} \label{sec:resonator_loihi}

\begin{figure}[ht]
  \centering
  \includegraphics[width=1.0\linewidth]{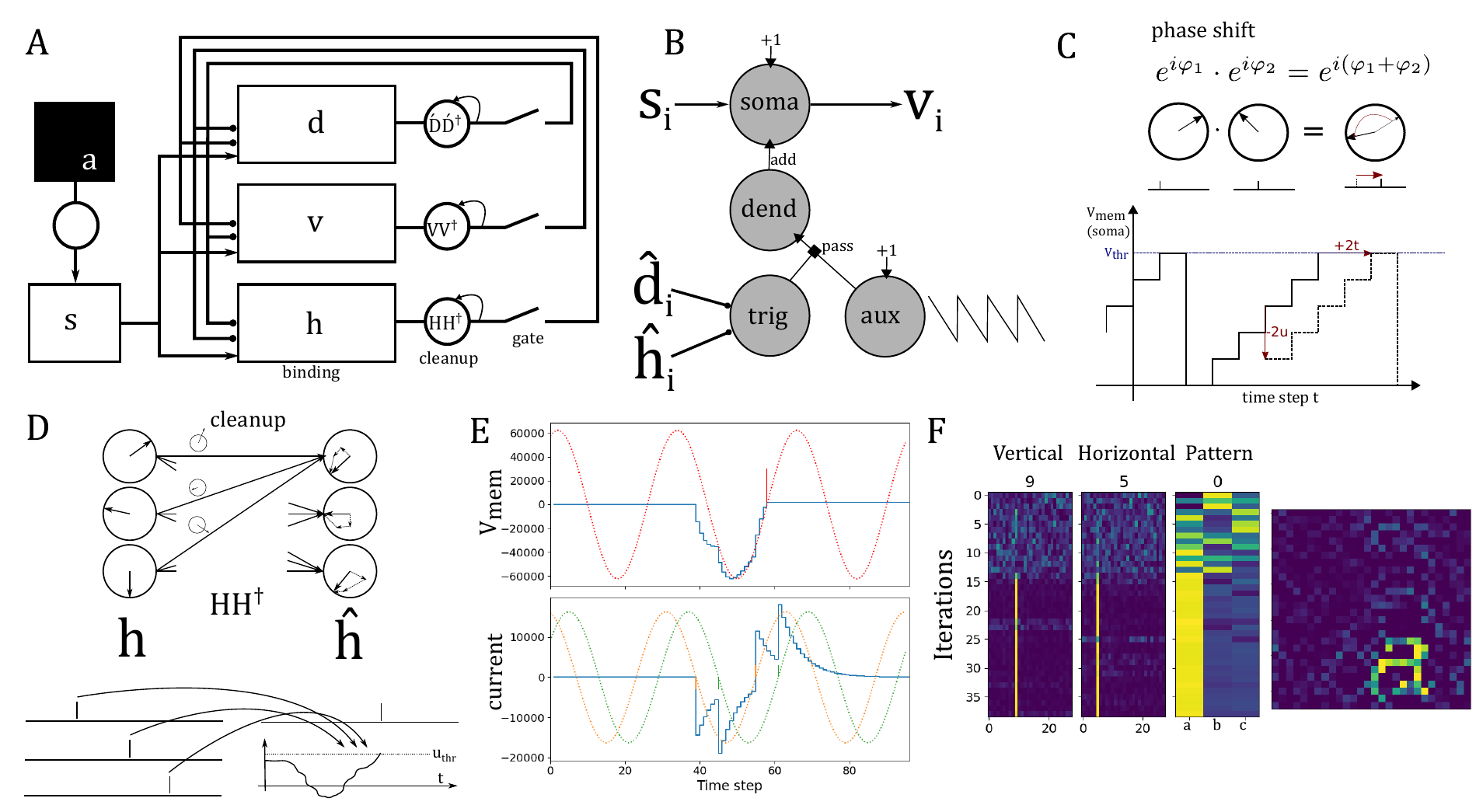}
  \caption{The resonator network on the Loihi neuromorphic hardware.
  \textbf{A.} Schematic of the resonator architecture implemented on Loihi.
  \textbf{B.} Implementation of the (un-)binding module as a 4-compartment neuron on Loihi. 
  \textbf{C.} Mechanism of the phase shift. 
  Here, the soma membrane potential is inhibited by 2, so it will reach the threshold two timesteps later. 
  The inset at the top shows the equation of complex multiplication, its phasor representation, and the corresponding spike timing in phasor I\&F neurons.
  \textbf{D.} Mechanism of the cleanup module. Top: Phasor representation of the complex matrix multiplication of h with the cleanup matrix $\mathbf{H} \mathbf{H}^\dagger$. 
  Bottom: The same mechanism with I\&F phasor neurons.
  \textbf{E.} Mechanism of the complex adder with I\&F phasor neurons. The neuron receives two inputs at different phases (orange and green). 
  The current gets integrated into the membrane potential, which approximates a sine wave (red). 
  In blue, the membrane potential and input current of the Loihi neuron are shown. 
  \textbf{F.} States of the resonator on Loihi over 40 iterations and reconstructed image from resonator states.
  }
  \label{fig:loihi_implementation}
\end{figure}

Finally, we implement the Resonator model for visual scene analysis with an efficient spike-timing code running on Intel's Loihi neuromorphic research chip \citep{davies2018loihi}. In this proof-of-concept, the resonator modules for translation ($\mathbf{\hat{h}}$, $\mathbf{\hat{v}}$) and pattern shape ($\mathbf{\hat{d}}$) are mapped to a spiking neural network (SNN; Fig. \ref{fig:loihi_implementation}A).
Specifically, in reference to a background oscillation, the spike times of integrate and fire (I\&F) neurons represent the phases of complex numbers in the VFA hypervectors \citep{frady_2019tpam} (Fig.~\ref{fig:loihi_implementation}B).
The Loihi chip has discrete-time dynamics, and the predefined cycle window is of length $T=16$ timesteps. 

The spiking neural network resonator network on Loihi consists of three factor modules that are connected recurrently. The spike generator converts the input vector $\mathbf{s}$ into spikes and transmits these spikes to the binding stage of each module (Fig.~\ref{fig:loihi_implementation}B). Each binding stage performs a neuron-wise complex phase shift based on its inputs (Fig.~\ref{fig:loihi_implementation}C), implementing the FHRR binding operation. This is computed using a multi-compartment neuron model available on Loihi that allows input spikes to the dendritic compartment to shift the output spike-timing by their own phase (see Methods).
The cleanup module (Fig.~\ref{fig:loihi_implementation}D) performs a matrix multiplication with the auto-associative matrix, e.g., $\mathbf{H}\mathbf{H}^\dagger$. 
This is computed using synaptic delay mechanisms and post-synaptic potentials that are configured to approximate a cycle of a sine wave, as described in \citep{frady_2019tpam} (Fig.~\ref{fig:loihi_implementation}E; see Methods).
An additional gate stage in each module controls the flow of spikes through the network, ensuring the network maintains the correct timing. 

\begin{figure}[ht]
  \centering
  \includegraphics[width=0.7\linewidth]{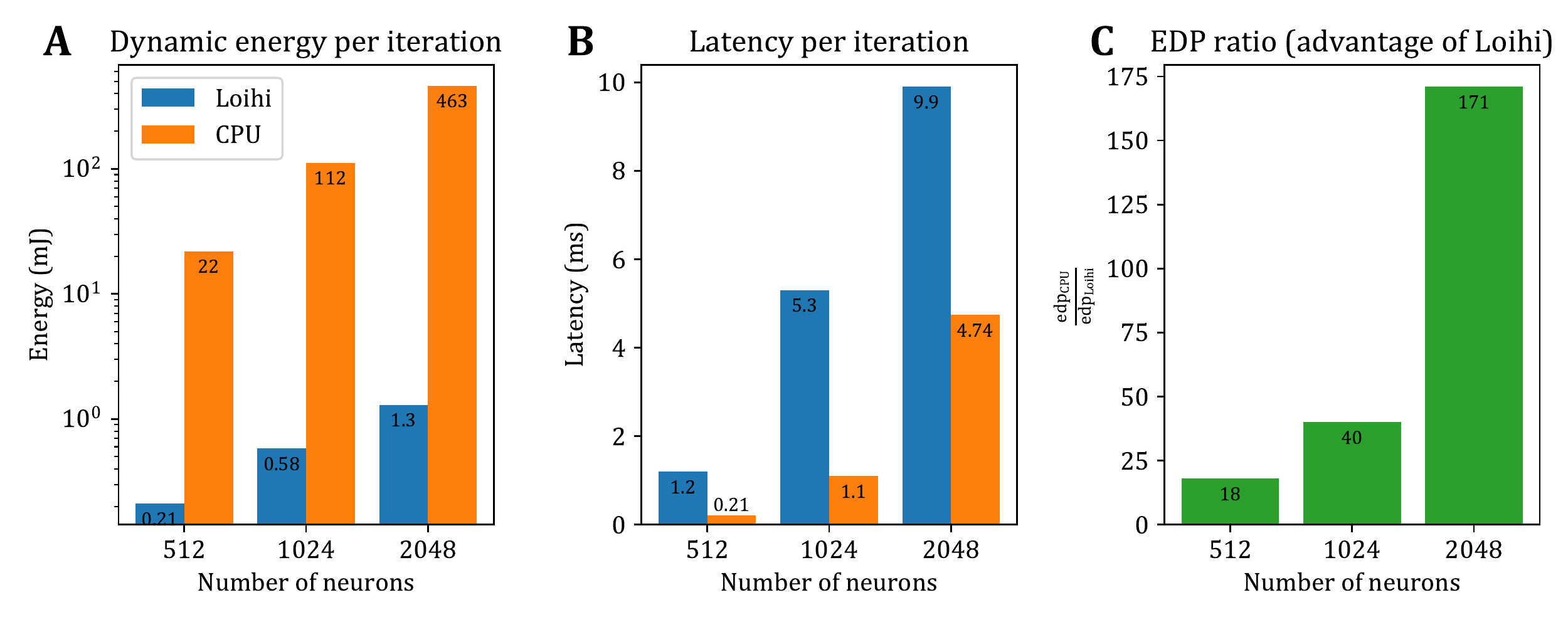}
  \caption{Delay and energy comparison. \textbf{A.} Logarithmic plot of the dynamic energy on Loihi and CPU.
  \textbf{B.} Time per iteration.
  \textbf{C.} EDP ratio.}
  \label{fig:power_comparison}
\end{figure}

Figure~\ref{fig:power_comparison} shows a comparison between Loihi and a CPU in terms of energy and latency. While Loihi is slower, it is orders of magnitude more energy efficient (Fig.~\ref{fig:power_comparison}A and B). For the largest network size, Loihi is 171 times more efficient in terms of energy-delay-product (EDP) (Fig.~\ref{fig:power_comparison}C) and scales better than the CPU for increasing network size, likely due to sparse, event-based matrix multiplication \citep{DaviesAdvancingLoihi2021}.

\section*{Discussion}

We have described a new network architecture for inference in generative models and its neuromorphic implementation with an efficient spike-timing code \citep{frady_2019tpam} on modern spiking neuromorphic hardware \citep{davies2018loihi}. This network architecture is validated on the problem of analyzing synthetic visual scenes, and our companion paper demonstrates an application to a real-world task, visual odometry in robotics~\citep{renner2022A}. 

Inference in generative models -- also known as ``analysis by synthesis'' -- is a powerful method for invariant pattern analysis, but it is infamously computationally expensive \citep{teh2003energy}. Our proposal to make it tractable hinges on four key ideas.
The first idea is to develop data encoding schemes and generative models that make binding the equivariant operator for a factor of variation, such as image translation. The second idea is to compute inference in the generative model by a resonator network that searches the solution space efficiently by computing in superposition \cite{frady_2020_resonator}. 
The third idea addresses the existence of multiple non-commutative factors of variation, like translation versus image rotation and scaling. Following a classical method for image registration, the Fourier-Mellin transform~\citep{casasent1976position,chen1994symmetric,reddy1996fft}, we propose a novel hierarchical resonator network, where binding is the equivariant operation for translation in one partition and for rotation and scaling in the other.  
The fourth idea is to use multi-compartment neuron models for implementing the network in a spike-timing code on neuromorphic hardware.

Avoiding the inefficiency of neuromorphic computing with rate-codes \citep{Davies_etal18}, our neuromorphic implementation employs a spike phase code, which outperforms an implementation on a CPU in efficiency. However, the implementation demonstrated here is rather a proof of concept. The model formulation is flexible and can be further optimized for a specific neuromorphic and other edge AI hardware implementation.

In artificial intelligence, there has been interest in interpretable factorized or ``disentangled'' representations \citep{bengio2013representation,higgins2017betavae}, with a particular focus on learning factorized representations from data \citep{kingma2013auto,higgins2017betavae, Tran_2017_CVPR,chau2020disentangling}. However, existing approaches rely upon generic neural network architectures without explicitly accommodating the computations necessary for factorization. It has also been noted that fully equivariant generalization in such learned representations is limited and ``brittle'' \citep{fil2021beta}, leading even to the conjecture that unsupervised learning of true disentangled representations is ``impossible'' \citep{locatello2019challenging,khemakhem2020variational} without an inductive bias, i.e., mechanisms for generalization that are engineered into the network (or into the “augmentation” of training data sets \cite{li2022neural}) rather than being learned.

Here, we focus on the fundamental computational principles of scene analysis -- i.e., factorization and inference -- rather than on learning per se. 
By approaching perception as a problem of inference in a generative model, we avoid the shortcomings of models that rely purely on learned representations, such as short-cut learning \citep{geirhos2020shortcut,eulig2021diagvib}, as well as lack of out-of-distribution generalization \citep{alcorn2019strike, wenzel2022assaying,frady2023learning} and extrapolation \citep{montero2020role,schott2021visual}. 
In addition, we also provide a meaningful definition of disentangled representations by directly linking disentanglement to vector factorization \citep{kim18disentangling}, an understanding that has been lacking so far \citep{fil2021beta}.

When evaluated on the analysis of simple synthetic scenes, the model succeeds most of the time but can also fail for several reasons, most of them common to other generative model approaches.
The primary reason for mistakes is spurious matches involving ambiguity through symmetries, e.g., `p' versus a rotated `d', or complex mixtures of generative factors. The spurious matches reconstruct the input with high quality, correctly capturing location and color but using shape primitives of the wrong class. 
To prevent the explanation of an object by a complex mixture of objects, we explored the addition of sparsity mechanisms to the resonator dynamics that encourage simple scene explanations, as well as further alignment and annealing procedures and observed improved performance (see Methods and Supplementary Table \ref{tab:ablation}). A better option in the future would be the concise inclusion of factor priors in the model.
Another cause for spurious matches that is harder to fix is mutual correlations between shape templates. 
While we do eliminate correlations between shape templates in their default pose, strong correlations between templates can still arise if their relative poses differ.
The earlier analysis of resonator networks \citep{kent_2020resonator} assumed the total absence of correlations between code vectors and, therefore, provided an upper bound on the complexity of a scene that a resonator of a given size can handle. According to the theory, the number of combinations that can be searched scales with the square of network size, i.e., $M_{max} \propto N^2$. 
For scenes with small objects that extend just a few pixels, the theory in \citep{kent_2020resonator} predicts that the dimension of the hypervectors has to grow roughly with the square root of the number of pixels.
Further, the number of hypervectors to be stored is only $P_x+P_y$ since the network architecture divides each pixel dimension into separate modules.

The presented results demonstrate that the combination of neuromorphic hardware and recent developments in VSA offers a tractable approach to scene analysis through a generative model. 
The approach with hierarchical resonator networks may extend beyond synthetic scenes used in our benchmark. 
For instance, one direction we are currently investigating is describing a larger variety of visual objects in the generative model using sparse features rather than full-letter templates.
The path towards neuromorphic analysis of naturalistic scenes, useful for embodied agents, will require further revising the generative model (\ref{eq:gen_synth}), such as adding priors, 3D shape models with 3D transforms \citep{chaudhuri2020learning}, and effects of occlusion \citep{huang2015efficient}.

Last but not least, the work also has implications for neuroscience. The feature binding problem in neuroscience 
\cite{feldman2013neural} is solved seamlessly by vector binding, which is at the core of this scene understanding model.
Our neuromorphic implementation of scene understanding makes predictions on the function of oscillations and spike phase locking that differ entirely from earlier proposals on how binding information might be represented by synchrony \cite{Gray_Singer89} or asynchrony \cite{nadasdy2010binding} in rhythmic spike codes.

\section*{Methods}

\subsection*{Details of simulation experiments}
Simulation experiments using the resonator network were implemented in Python using Numpy and PyTorch. The vector dimensions used for the benchmarking experiments are $N=10,000$ for the translation task, $N=16,384$ for the Cartesian module, and $N=22,680$ for the log-polar module of the hierarchical resonator in the rigid transform task.

The resonator network is initialized to either a random state or each factor is initialized to the mean of all its codebook vectors (leading to slightly more reliable performance).

The full dynamic equations for the hierarchical resonator network:
\begin{align}
\begin{split}
    \mathbf{\hat{c}}(t+1) &= f \Big( \mathbf{\acute{C}}\mathbf{\acute{C}}^\dagger \Big( \mathbf{s} \odot \mathbf{\hat{h}}^*(t) \odot \mathbf{\hat{v}}^*(t) \odot \big[ \mathbf{\Lambda}^{-1} \big( \mathbf{\hat{r}}(t) \odot \mathbf{\hat{m}}(t) \odot \mathbf{\hat{d}}(t) \big) \big]^* \Big) \Big), \\
    \mathbf{\hat{h}}(t+1) &= f \Big( \mathbf{H}\mathbf{H}^\dagger \Big( \mathbf{s} \odot \mathbf{\hat{c}}^*(t) \odot \mathbf{\hat{v}}^*(t) \odot \big[ \mathbf{\Lambda}^{-1} \big( \mathbf{\hat{r}}(t) \odot \mathbf{\hat{m}}(t) \odot \mathbf{\hat{d}}(t) \big) \big]^* \Big) \Big), \\
    \mathbf{\hat{v}}(t+1) &= f \Big( \mathbf{V}\mathbf{V}^\dagger \Big( \mathbf{s} \odot \mathbf{\hat{c}}^*(t) \odot \mathbf{\hat{h}}^*(t) \odot \big[ \mathbf{\Lambda}^{-1} \big( \mathbf{\hat{r}}(t) \odot \mathbf{\hat{m}}(t) \odot \mathbf{\hat{d}}(t) \big) \big]^* \Big) \Big), \\
    \mathbf{\hat{d}}(t+1) &= f \Big( \mathbf{\acute{D}_L} \mathbf{\acute{D}_L}^\dagger \Big( \big[ \mathbf{\Lambda} \big( \mathbf{s} \odot \mathbf{\hat{c}}^*(t) \odot \mathbf{\hat{h}}^*(t) \odot \mathbf{\hat{v}}^*(t)  \big) \big] \odot \mathbf{\hat{r}}^*(t) \odot \mathbf{\hat{m}}^*(t) \Big) \Big), \\
    \mathbf{\hat{r}}(t+1) &= f \Big( \mathbf{R} \mathbf{R}^\dagger \Big( \big[ \mathbf{\Lambda} \big( \mathbf{s} \odot \mathbf{\hat{c}}^*(t) \odot \mathbf{\hat{h}}^*(t) \odot \mathbf{\hat{v}}^*(t)  \big) \big] \odot \mathbf{\hat{d}}^*(t) \odot \mathbf{\hat{m}}^*(t) \Big) \Big), \\
    \mathbf{\hat{m}}(t+1) &= f \Big( \mathbf{M} \mathbf{M}^\dagger \Big( \big[ \mathbf{\Lambda} \big( \mathbf{s} \odot \mathbf{\hat{c}}^*(t) \odot \mathbf{\hat{h}}^*(t) \odot \mathbf{\hat{v}}^*(t)  \big) \big] \odot \mathbf{\hat{d}}^*(t) \odot \mathbf{\hat{r}}^*(t) \Big) \Big). 
    \label{eq:hierarch_res}
\end{split}
\end{align}
Note that the repeated terms, $\mathbf{\Lambda} \big( \mathbf{s} \odot \mathbf{\hat{c}}^*(t) \odot \mathbf{\hat{h}}^*(t) \odot \mathbf{\hat{v}}^*(t)  \big)$ and $\mathbf{\Lambda}^{-1} \big( \mathbf{\hat{r}}(t) \odot \mathbf{\hat{m}}(t) \odot \mathbf{\hat{d}}(t) \big)$, are simplified into resonator bridge modules $\mathbf{\hat{l}}$ (\ref{eq:bridgeeq1}) and $\mathbf{\hat{p}}$ (\ref{eq:bridgeeq2}).

The hierarchical resonator network is a combination of the resonator network for translation (\ref{eqn:resonator_transla}) and the network for rotation/scaling (\ref{eqn:lp_resonator}).
The \emph{log-polar partition}, the right column of modules in Fig.~\ref{fig:hier_res}A, contains the ``top-down'' bridge module (\ref{eq:bridgeeq1}) and the modules from (\ref{eqn:lp_resonator}).
The top-down bridge module (\ref{eq:bridgeeq1}) produces $\mathbf{\hat{l}}$, the estimate of a rotated and scaled shape transformed to Cartesian coordinates.
The \emph{Cartesian partition}, the left column of modules in Fig.~\ref{fig:hier_res}A, contains the modules from (\ref{eqn:resonator_transla}), but the module with output $\mathbf{\hat{d}}$ replaced by the ``bottom-up'' bridge module (\ref{eq:bridgeeq2}). The previous input from $\mathbf{\hat{d}}$ is replaced by the top-down signal $\mathbf{\hat{l}}$. 
The bottom-up bridge unit (\ref{eq:bridgeeq2}) produces $\mathbf{\hat{p}}$, which transforms a centered version of the input into log-polar coordinates. This, in effect, replaces the input for (\ref{eqn:lp_resonator}).

We call the architecture in Fig.~\ref{fig:hier_res}A the \emph{hierarchical resonator network} because the bidirectionally connected partitions assume different hierarchy levels by the definition of Felleman and Van Essen \citep{felleman1991distributed}. The Cartesian partition receives direct input according to a lower level, while the log-polar partition is one removed from the sensory input according to a higher level. The log-polar partition also holds the discrete shape templates in memory, the arguably most abstract aspect of the image components. 

The codebooks $\mathbf{\acute{D}_L} \in \mathbb{C}^{N \times 26}$ and $\mathbf{\acute{C}} \in \mathbb{C}^{N \times 7}$ are formed from whitened templates projected into the high-dimensional vector space. Specifically, the templates are transformed into log-polar coordinates and decomposed by the singular value decomposition, i.e., $\mathbf{P_L} = \mathbf{L} \mathbf{P} = \mathbf{U_L} \mathbf{\Sigma_L} \mathbf{V_L}$. 
The whitened templates are given by $\mathbf{\acute{P}_L} = \mathbf{U_L} \mathbf{V_L}$, which is then projected into the high-dimensional space by $\mathbf{\acute{D}_L} = \mathbf{\Phi_L} \mathbf{\acute{P}}$. A similar whitening procedure is applied for the colors to produce $\mathbf{\acute{C}}$, but not for the other codebooks.
In the resonator for the translation task, the letter template images are aligned for the whitening procedure to ensure the whitening removes the correlation at the most relevant shift of the letters. For instance, for the whitening of each letter image, all other letters are aligned (pairwise to the respective letter) by finding the best overlap using image registration by phase correlation in the 2d Fourier domain \citep{casasent1976position,chen1994symmetric,reddy1996fft}.
Then, whitening is performed over the letter image and all pairwise aligned other images. The respective whitened image is added to the codebook, and all other images are discarded. This is repeated for all letter images.
For the hierarchical resonator networks, this alignment-whitening procedure is omitted as it is impossible to capture all relevant correlations and therefore, performance would not improve. For the hierarchical resonator network, letters are just aligned before adding them to the codebook.

In the hierarchical resonator network, rotation as an operation must have the correct topology for its representation -- i.e., rotation by $360^o$ results in the same image. This is done by ensuring that the representation vector $\mathbf{r}$ also has circular topology. The circular topology is encoded by defining a periodic kernel for the representation $\mathbf{r}$, where the random phases of the elements of $\mathbf{r}$ are sampled from a discrete probability distribution \citep{frady2021VFA}. Specifically, the phase circle is divided into $L_r$ discrete samples and each element of $\mathbf{r}$ is one of these samples, $r_i \in \{ e^{i 2 \pi k / L_r} \  \forall k \in \{1, ..., L_r\} \}$. 

Note that our definition of binding acting as the equivariant operation to image translation (\ref{imagetrans}) requires some leniency when considering the edges of the image. The definition strictly holds if we assume that the image has a toroidal topology, and the position vectors $\mathbf{h}$ and $\mathbf{v}$ also have the correct topological structure as described above. 
Also note that the complex exponential is multi-valued, i.e. $(e^{i\varphi_j})^x = e^{x(i\varphi_j + 2 \pi n)} \, \forall n \in \mathbb{Z}$, but we define the operation as returning only the principal value with $n=0$.

Beyond the dynamics described in equation (\ref{eq:hierarch_res}), we include some modifications to improve performance.
One modification is hysteresis in the update dynamics, where some fraction of the past state is included in the next update, i.e.:
\begin{equation}
    \mathbf{\hat{d}}(t+1) = (1 - \gamma) \mathbf{\hat{d}}(t) + \gamma f(\acute{\mathbf{D}} \acute{\mathbf{D^\dagger}} (\mathbf{s} \odot \mathbf{\hat{h}}^*(t) \odot \mathbf{\hat{v}}^*(t)),
    \label{eq:hysteresis}
\end{equation}
with $\gamma$ controlling the rate of the hysteresis.

Another modification is adding a non-linearity, such as a ReLU, polynomial exponent, or softmax function, to encourage sparsity. Additionally, we incorporated external noise ($\eta$) to reduce the stability of spurious local minima: 
\begin{equation}
    \mathbf{\hat{d}}(t+1) =  f(\acute{\mathbf{D}} p( \acute{\mathbf{D^\dagger}} (\mathbf{s} \odot \mathbf{\hat{h}}^*(t) \odot \mathbf{\hat{v}}^*(t))) + \eta.
    \label{cleanup2}
\end{equation}
Our experiments show that a combination of a ReLU and polynomial exponent $p(\mathbf{x}) = \mathbf{ReLU(x)}^k$ performs best. The ReLU serves two purposes: It avoids negative factor pairs and acts as a threshold to improve the cleanup. Note that without ReLU, the resonator network can converge to a solution where an even number of factors are negative, a problem that can be fixed by taking absolute values as the final outputs. The $k$ parameter controls the amount of superposition of different possible solutions, i.e., the sparsity. It can be set to values below 1 to weaken the cleanup or to values well above 1 to achieve a stricter cleanup (like an argmax or winner-take-all for high $k$s). In the case of the translation resonator, $k=1$ proved to be sufficient, while in the hierarchical resonator, at least one factor (we chose the angle factor) needs a larger $k$, typically above 3.
Furthermore, after each iteration, the resonator states (with the exception of the pattern state ($\mathbf{d}$)) are phasor projected ($f$), i.e., magnitudes of all vector elements are set to 1. 
Additionally, complex Gaussian distributed noise is added to the state ($\eta \in \mathbb{C}^N \sim \mathcal{N}(0,\sigma)$) with $\sigma=1$ in the resonator for the translation task and $\sigma=0$ in the hierarchical resonator's log-polar module. In the last 2 iterations of each pass, however, the noise is turned off to get a cleaner readout.

\subsection*{Details of performance benchmarking}
To benchmark the model, images with a given number of letters are created. Letters, locations, and rotations are chosen randomly. For the translation task, letters are chosen randomly from all 26 letters of the English alphabet; for the rotation task, from the first 10 letters of the alphabet. We use the font ‘TlwgTypewriter-Oblique’ at a font size of 26 in an image of 64x64 pixels. The letters are shifted using scipy.ndimage.shift by a random floating point number uniformly distributed between -19 and +19, i.e., with a margin of 13 pixels from the border to avoid cropping of letters. To avoid overlaps between letters, after adding a letter to an image, the new image is compared with the old image. If at least one pixel overlaps, the random choice of vertical and horizontal translation of the newly added letter is repeated. If no non-overlapping image has been found after 20 repetitions of this process, the image is used as it is in the 20th repetition (with the overlap). Note that letters can still look like they overlap if they have colors that are non-overlapping, such as red and green.

The complexity of the scene analysis problem can be quantified by the total combinations that the system must search over. 
In the benchmark, there are 26 letters and 7 colors and a range of translations covering 39x39 pixels, giving a combination space of 276,822. This amount is much less than the operational capacity (which for $N=12,000$ is about $100$ million) \citep{frady_2020_resonator} where the resonator network is expected to have nearly $100\%$ accuracy. 
Note that with 6 factors of variation, the combinatorial space has expanded to over 100 million combinations the network must search through, meaning the problem is much more challenging.
Importantly, note that the operational capacity in a resonator was measured for uncorrelated vectors, and performance on scene analysis will be different due to correlations between the generative objects. The whitening removes correlations when letters are in their default position but not when they are in arbitrary relative positions. 

To benchmark the different model variants, we report an accuracy measure based on the percentage of correct classifications of the letter identity. We chose this measure as it is straightforward to calculate and compare and because getting the letter identity correct usually also means that the other factors are estimated well (the opposite is not true).
The letter output of the resonator is the argmax of the state readout (product of the letter factor codebook and the factor state).

For the tasks with several letters in the input image, a classification is counted as correct if the letter output corresponds to any of the letters in the image. One instance of the correctly guessed letter is then removed from the list so that the next pass has to guess one of the remaining letters correctly to count as a correct classification.

To find the best parameters for the different model variants, we perform a hyperparameter search. Starting from a manually tuned network, we perform a sweep for each parameter separately. Parameters that are optimized in this manner are the noise variance ($\sigma$), the state update ratios for each set of states separately ($\gamma$), and the polynomial exponents ($k$).
The hyperparameter search is performed with a different random seed, i.e., a different dataset and a different random codebook than the test seed that is only used to test the network on a given number of samples.

\subsection*{Details of hardware implementation}

We implement a smaller model that solves a 28x28x3 factorization task on Intel's neuromorphic research chip Loihi~\citep{davies2018loihi}. The smaller version is implemented on the USB form factor system ``Kapoho Bay," which features two Loihi chips with a total of 256 neuro cores in which a total of 262,144 neurons and up to 260 million synapses can be implemented. The embedded x86 processors are used for monitoring and sending input spikes.

Using phasors in a spiking network makes multiplication and addition of complex vectors available on the hardware.
Phasors can be implemented with resonate and fire neurons or, by adding certain constraints, with common integrate and fire neurons \citep{frady_2019tpam}.
In this work, we use I\&F neurons with a single spike per cycle of T=16 timesteps, which allows us to represent complex numbers of unit magnitude and discrete (4-bit) phase. Each complex element of a hyperdimensional vector is implemented on the hardware as a neuron. 

The binding operation in FHRR is the Hadamard product (elementwise multiplication) between complex vectors.
The multiplication of unit complex numbers is just the addition of their phases. We represent the phases of complex numbers as spike timing, and thus, binding is a shift of spike times.
To achieve the correct shift of spike-timing, we designed a 4-compartment I\&F neuron compatible with Loihi 1.
The structure of this neuron is shown in Figure \ref{fig:loihi_implementation}B and the mechanism is illustrated in Figure \ref{fig:loihi_implementation}C.

Two of the compartments, ``aux'' and ``soma'', act like clocks with a cycle time of $T$ by simply integrating a constant value (e.g., 1) and resetting when they reach a threshold (e.g., 16 to achieve a cycle of T=16 timesteps). The first input ($s$) to the soma compartment resets the soma's clock. The second input spike to the ``trig'' compartment opens a gate from the ``aux'' to the ``dend'' compartment. When the gate is opened, the aux state, which corresponds to the (negative) phase of the second input spike, is added to the soma state, delaying the soma from reaching the spike threshold and thus shifting the timing of the output spike (as illustrated in Fig.~\ref{fig:loihi_implementation}C). Note that this multi-compartment implementation of phase addition is specific to the Loihi 1 hardware; the following equations describe the binding module using a discrete-time I\&F mechanism independent of the hardware for a single neuron (vector element):
\begin{equation}
\begin{aligned}
    v(t+1) &= v(t) + 1 - v(t) s_0(t) - c(t)s_1(t)\\
    c(t+1) &= c(t) + 1 \\
\end{aligned}
\end{equation}
The membrane potential $v(t)$ corresponds to the state of the ``soma'' compartment, and the clock $c(t)$ corresponds to the state of the ``aux'' compartment. Both states are reset to 0 when they reach a threshold $v_{Thr}=T$, and the ``soma'' compartment sends an output spike. 
The $s_0(t)$ and $s_1(t)$ represent input spike trains, with a value of $1$ at the spike times and $0$ everywhere else. 
The $s_0$ signal determines each neuron's reference phase and the $s_1$ signal represents the inputs that shift the phase for binding.
This mechanism takes maximally one cycle per additional input spike (which becomes relevant when more than two vectors are bound together) before the neuron settles to the correct output phase. The output spike time can be converted to the corresponding complex number by $e^{i 2 \pi \frac{t_{sout} \mod T}{T}}$.

Note that there are two versions of the binding circuit, one for binding and one for unbinding. Unbinding in FHRR is elementwise multiplication with the complex conjugate  (denoted by $*$ in the equations), i.e., phases are subtracted instead of added.
In the binding circuit, the `aux' compartment counts down from 0 to -T; in the unbinding circuit, it counts up from -T to 0. 

After the binding module, spikes are sent through a cleanup module, which performs a matrix multiplication with the auto-associative matrix, e.g., $HH^\dagger$. 
As described in Fig. \ref{fig:loihi_implementation}, each spike through the clean-up matrix elicits a negative and positive current impulse. 
The impulse is delayed using Loihi's synaptic delay settings, and the delay is based on the phase of the complex weight as described in \citep{frady_2019tpam}. 
This results in an approximation of sine-wave oscillations that occur in each I\&F neuron's membrane potential, which, when summed together, implement the complex dot product. 

To coordinate the timing of the network, there is a gate after the cleanup module. 
The gate module consists of an array of I\&F neurons of length $N$ that are connected in a 1:1 manner from the cleanup module to the binding module.
The neurons are inhibited by default and only allow spikes to relay when also depolarized by control input.
The synaptic delays of the gate are adjusted to ensure that the binding module and the cleanup module remain synchronized. 
Further, the gate allows for the cleanup module to iterate for three cycles before forwarding spikes to the binding module. 
The gate also ensures that only one spike per cycle is routed to the binding module.

In order to reach better convergence, the cleanup module has a recurrent 1:1 connectivity, i.e., each neuron is connected back to itself with a delay of one cycle T. This leads to slower evolution of the resonator state over iterations, akin to the hysteresis procedure (\ref{eq:hysteresis}). 

Because fanout (the number of connections allowed to leave a neuron) on the chip is limited per core, we distributed neurons on several cores over the chip and pruned half of the synapses with the lowest weights from the cleanup matrix. The dense cleanup matrix is the most limiting component of the architecture. An alternative would be to split the cleanup matrix into its two components, greatly reducing the number of synapses in most cases. However, the layer connecting the two matrix multiplications cannot be represented with a spiking phasor with unit magnitude. Graded spikes on the next version of Loihi could be used to enable phasors with a magnitude.

Power measurements for the Loihi 1 chip were obtained remotely using NxSDK version 0.9.9 on the Nahuku32 board ncl-ghrd-01. The Loihi board is interfaced to a system with an Intel Xeon CPU E5-2670 @ 2.60GHz and 128GiB of RAM running Ubuntu 20.04.4 LTS. Intel Labs provided both software API and hardware. All probes, including the output probes, except the energy probes, were deactivated. Energy was averaged over 20 resonator iterations.
Measurements for the CPU are obtained with Intel SoC Watch on a system with an Intel Core i9-7920X CPU @ 2.9GHz and 128GiB of RAM running Ubuntu 20.04 LTS. Simulations were run with Python 3.8.10 and NumPy 1.23.1. The process was constrained to use 12 threads since we found this to provide the best energy-delay-product measurement. The energy of the DRAM was not included. The latency and energy were averaged over 10000 iterations. Dynamic energy was measured by subtracting the static energy that is used when running the system without the load for the same amount of time. So, dynamic energy is the energy associated with the computation and excludes leakage energy.
The comparison uses cleanup with a full NxN auto-associative matrix on both Loihi and the CPU. For small numbers of symbols, it is computationally advantageous to first multiply with the decoding matrix (e.g., $H^\dagger$) and then with the encoding matrix ($H$). The intermediate result can, however, not be represented in a phasor encoding without amplitudes. We are working on an encoding that can represent amplitudes on Loihi 2, the next generation of the Loihi chip, which will improve the scalability of the architecture.

\section*{Data availability}
The synthetic images of letters used in the experiments can be recreated with the provided code on CodeOcean~\cite{renner_2024_codeocean_understanding}.

\section*{Code availability}
A notebook to demonstrate the resonator~\cite{frady_code} is available at \href{https://doi.org/10.5281/zenodo.10810900}{https://doi.org/10.5281/zenodo.10810900}.

The source code of the hierarchical resonator in PyTorch for benchmarking~\cite{renner_2024_codeocean_understanding} is available on CodeOcean at \href{https://doi.org/10.24433/CO.1543398.v1}{https://doi.org/10.24433/CO.1543398.v1}.

\section*{Acknowledgments}
A.R. discloses support for the research of this work from Accenture Labs, the University of Zurich postdoc grant [FK-21-136], and the VolkswagenStiftung [CLAM 9C854].
Y.S. and A.R. disclose support for the research of this work from the Swiss National Science Foundation (SNSF) [ELMA PZOOP2 168183].
F.T.S. discloses support for the research of this work from NIH [1R01EB026955-01] and NSF [IIS2211386].
We thank Intel Neuromorphic Computing Lab for providing access to the Loihi hardware and related software. The authors thank Elvin Hajizada for running the CPU power measurements.

\section*{Author Contributions Statement}
A.R., L.S., A.D., G.I., B.A.O., Y.S., F.T.S., and E.P.F. contributed to the writing and editing of the manuscript. A.R., E.P.F., and F.T.S. conceptualized the project. A.R. and E.P.F. developed the model and performed the simulation experiments and analysis. L.S. carried out the control experiments with YOLO (supplementary).

\newpage

\section*{Supplementary Material}

\subsection*{Notation and symbols reference}
\begin{table}[h!]
\centering
\resizebox{\linewidth}{!}{
\begin{tabular}{|l|l || l|l || l|l|}
\multicolumn{2}{l}{\textbf{Notation}} & \multicolumn{2}{l}{\textbf{Factors}} & 
\multicolumn{2}{l}{\textbf{Transforms}}\\
\hline
$\mathbf{\acute{A}}$ & Whitened codebook & $\mathbf{c}$ / $\mathbf{\acute{C}}$ & Color & $\mathbf{\Phi}$, $\mathbf{\Phi_P}$, $\mathbf{\Phi_L}$, $\mathbf{G}$ & VSA codebooks \\
$\mathbf{\hat{a}}$ & Factor estimate  & $\mathbf{h}$ / $\mathbf{H}$ & Horizontal position & $\mathbf{L}$ ($\mathbf{\Lambda})$ & Log-pol transf. (VSA space) \\
$\mathbf{A} ^\dagger$ & Complex conjugate transpose & $\mathbf{v}$ / $\mathbf{V}$ & Vertical position &
$\mathbf{s} = \mathbf{\Phi} \mathbf{I}$ & Input image \\
$\mathbf{a}^*$ & Complex conjugate & $\mathbf{r}$ / $\mathbf{R}$ & Rotation & 
$\mathbf{D} = \mathbf{\Phi_P} \mathbf{P}$ & Template images \\
$\mathbf{a} \odot \mathbf{b}$ & Binding (Hadamard Product) & $\mathbf{m}$ / $\mathbf{M}$ & Scale factor & 
$\mathbf{D_L} = \mathbf{\Lambda} \mathbf{D} = \mathbf{\Phi_L} \mathbf{L} \mathbf{P}$ & Log-polar templates \\
$\mathbf{a} + \mathbf{b}$ & Summation or superposition & $\mathbf{d}$ / $\mathbf{\acute{D}}$ / $\mathbf{\acute{D_L}}$ & Centered template & 
$\mathbf{C} = \mathbf{G} \mathbf{B}$ & Generative Colors \\
$\mathbf{a}^{-1}$ & Inverse & & & &\\
$\Re(a)$ & Real part of complex number & & & & \\
\hline
\end{tabular}
}
\caption{Notation and symbols reference}
\label{tab:notation_table}
\end{table}

\subsection*{Analysis of multiple objects} 
\begin{figure}[!htb]
  \centering
  \includegraphics[width=0.6\linewidth]{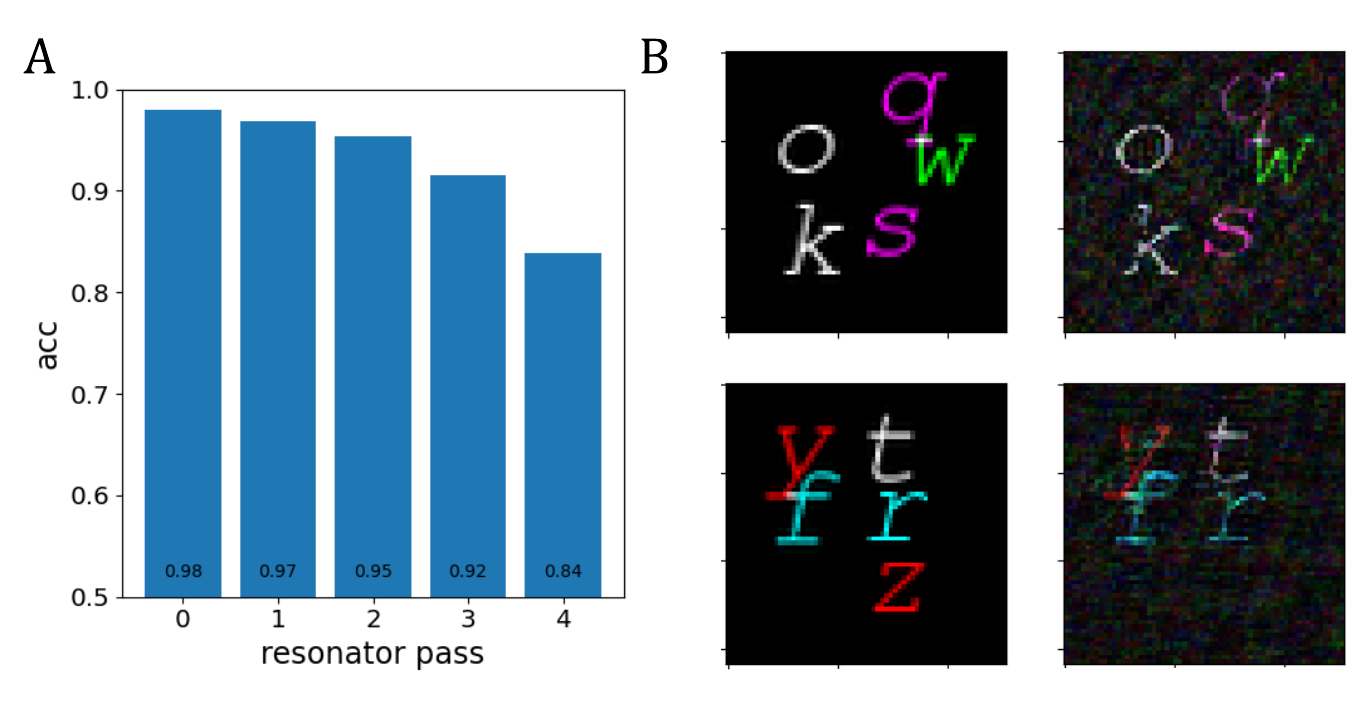}
  \caption{
  Benchmarking of the resonator network for the letter translation task with 5 random letters, and 1000 random samples. 
  \textbf{A.} Letter classification accuracy for each of the 5 passes.
  \textbf{B.} Two random example input images (left) and the reconstruction from the final resonator states (right), by adding all final states together. All letters were reconstructed correctly, apart from the red 'z' in the second image.
  }
  \label{fig:multiletter}
\end{figure}

Supplementary Fig. \ref{fig:multiletter} shows results for the resonator solving the letter translation task with five letters. The performance in the first pass is almost the same as for the translation task with a single letter. Performance for the following passes decreases slightly.  

\subsection*{Multi-headed resonator} 
\begin{figure}[!htb]
  \centering
  \includegraphics[width=0.7\linewidth]{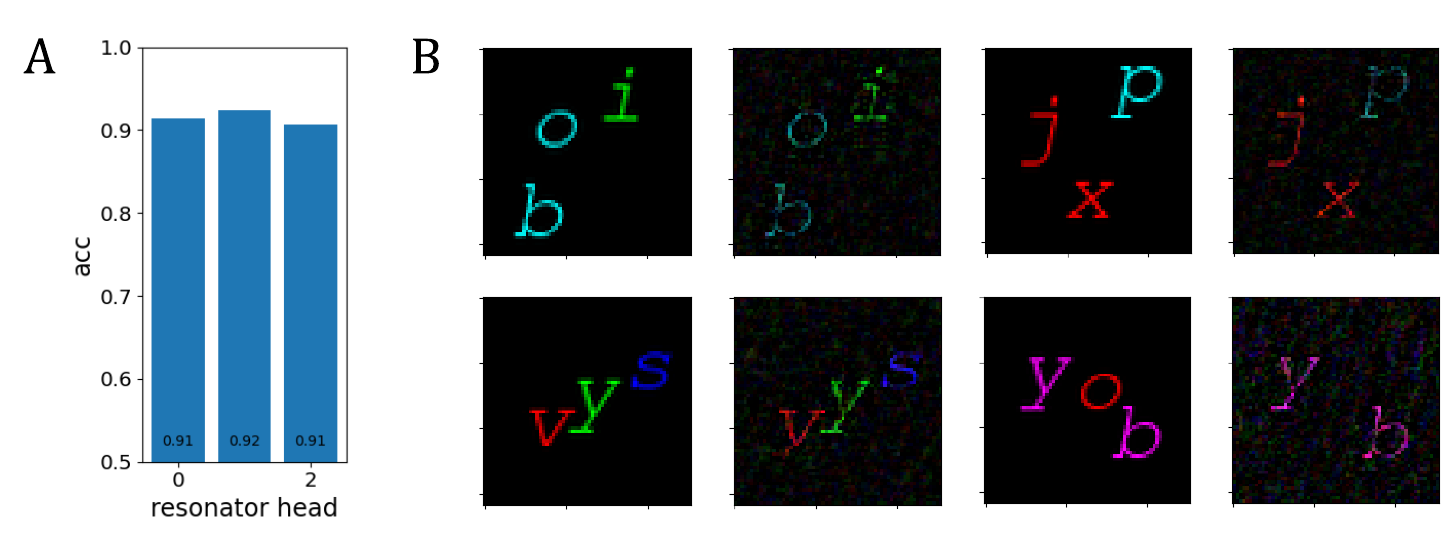}
  \caption{
    Benchmarking of the multi-headed resonator network for the letter translation task with 3 random letters, 500 random samples. 
  \textbf{A.} Letter classification accuracy for each of the 3 heads.
  \textbf{B.} Four random example input images and the reconstruction from the final resonator states. Bottom-right shows a mistake example that misses the red `o'.
  }
  \label{fig:multihead}
\end{figure}

Instead of running several passes of the resonator and explaining-away the result after each pass, we propose the \emph{multi-headed resonator}. The multi-headed resonator explains several letters in parallel. It consists of several identical copies of the resonator that run in parallel and compete to explain parts of the input image. The same explaining-away procedure is utilized but after each iteration instead of at the end of the pass. Whenever one head has found and is close to converging to one of the letters, this letter is subtracted from the input and, therefore, becomes invisible to the other heads. Supplementary Fig. \ref{fig:multihead} shows the results for the 3-headed resonator solving the translation task with three letters.

\subsection*{Explaining away procedures} 
We explore explaining away both in the image space and in the latent (VSA) space.
In the VSA space, the product of all resonator states (i.e., all factors bound together) is subtracted from the encoded input before it is sent to the next pass or used by the other heads.
For explaining-away in the image space, the resonator network state is used to reconstruct the image. This image is then max normalized and all values below a threshold (0.1) are set to 0. 
The reconstruction image is rescaled and then subtracted from the input image to remove all correlation. The result is thresholded again to get rid of values below 0. The resulting residual image is then encoded into a VSA vector and used as input.

The subtraction in the latent VSA space is computationally cheaper as the image does not have to be reconstructed, but so far comes with a drop in performance (of 10-15\%) as the states are less clean and additionally cannot fully explain away the input image due to the whitening. One could experiment with additional, stronger (higher k) cleanup with the non-whitened codebook for this purpose.

\subsection*{Ablation experiments} 
We test different variants of the resonator model to determine the importance of the algorithmic components, see Supplementary Table \ref{tab:ablation}.

\begin{table}[t]
\centering
\begin{tabular}{@{}lr@{}}
\toprule
model version & accuracy (\%) \\
\midrule
full model & 98.4 \\ 
without noise & 96.0 \\
without aligned whitening & 93.8 \\ 
without ReLU ($k=1$, $N=30K$) & 92.6\\
without ReLU ($k=2$, $N=10K$) & 92.1 \\
\bottomrule
\end{tabular}
\caption{\textbf{Testing of different model variants of the resonator on the letter translation task.}}
\label{tab:ablation}
\end{table}

\subsection*{Comparison with supervised learning approaches}

We explored the use of deep neural network architectures trained with supervised learning on our task in additional experiments. 
As described in our previous experiments \citep{frady2023learning} and from several other studies \citep{locatello2019challenging}, systems trained to learn the factors of variation from data fall short of ``out-of-distribution'' generalization. 
This means that if particular combinations of generative factors are not present in the training data, then the system will fail to recognize this factor conjunction during testing.
However, network architectures with a built-in inductive bias can overcome this.

To verify these results, we performed additional experiments testing out-of-distribution generalization in different types of neural networks.
In our previous work \citep{frady2023learning}, we showed that deep learning networks without any inductive bias fail to generalize to a combination of shape and translation that was not present in the training data, such as a 7 being shown in the bottom left corner of the image. 

Convolutional neural networks, through their architecture, include an inductive bias towards translation invariance, but surprisingly often do not achieve translation invariance \citep{azulay2018deep}. 
In our experiments, we see that a shallow 3-layer CNN fails in out-of-distribution generalization (acc. $<25\%$).
However, we tested a much larger CNN architecture, ResNet18, as well as YOLO (v3)  \cite{redmon2016you}, which can indeed perform out-of-distribution generalization with high accuracy ($>90\%$). We tested this by excluding 30\% of the letter translation in one direction from the training data.

The YOLO (v3) architecture includes an inductive bias for translation by generating bounding boxes around objects before classification. 
However, there is not an inductive bias for object rotation, and we again tested YOLO's out-of-distribution generalization for object rotations. 
When trained with a dataset that leaves out 30\% (the first 54°) of the rotations, YOLO classification accuracy drops below $50\%$ on the out-of-distribution samples. 

In conclusion, our approach may inspire novel solutions to the following challenges:
1. Generalization over new factors of variation: In CNNs, the factor over which generalization is possible (i.e., translation) is baked into the system as an inductive bias. In the resonator, factors can be added by the binding operation if the data embedding is constructed or learned accordingly.
2. To achieve invariance or equivariance, through supervised or unsupervised methods, a large amount of training data is required due to the mentioned lack of generalization.
3. Catastrophic forgetting and continual learning: New classes cannot easily be added without retraining. In the case of the resonator, only the respective factor where a class is added needs to be adjusted.
4. Data representations in many forms of deep networks require slot-based or space-based feature binding. Thus, they cannot easily represent several compositional objects in a single memory slot without running into a binding problem (mixing up features of the stored objects).  

\end{document}